\documentclass[runningheads]{llncs}
\usepackage[T1]{fontenc}
\usepackage{subfig}
\usepackage{graphicx}
\usepackage{hyperref}
\usepackage{caption}
\usepackage{adjustbox}
\usepackage{array}
\usepackage{booktabs}
\usepackage{color}

\newcommand{\citep}{\cite}
\newcommand{\citet}{\cite}

\begin{document}
\title{Stochastic Parrots Looking for Stochastic Parrots:\\
LLMs are Easy to Fine-Tune and Hard to Detect\\
with other LLMs}
\titlerunning{Stochastic Parrots Hide-and-Seek}
\author{Da Silva Gameiro Henrique\inst{1} \and
Andrei Kucharavy\inst{1, 2,}\thanks{Contact author}\and
Rachid Guerraoui\inst{1}}

% Andrei Kucharavy\inst{2, 3}\orcidID{0000-0003-0429-8644} \and
% Rachid Guerraoui\inst{2}\orcidID{0000-0002-4794-8902}
  
%
\authorrunning{H. Da Silva Gameiro et al.}
\institute{IC School EPFL, Switzerland\\
\email{\{firstname.name\}@epfl.ch}\\ \hspace{1cm}
\and
Now at HES-SO Valais-Wallis, Switzerland\\
\email{\{firstname.name\}@hevs.ch}\\ \hspace{1cm}
}

\maketitle              % typeset the header of the contribution
\begin{abstract}

The self-attention revolution allowed generative language models to scale and achieve increasingly impressive abilities. Such models - commonly referred to as Large Language Models (LLMs) - have recently gained prominence with the general public, thanks to conversational fine-tuning, putting their behavior in line with public expectations regarding AI. This prominence amplified prior concerns regarding the misuse of LLMs and led to the emergence of numerous tools to detect LLMs in the wild.

Unfortunately, most such tools are critically flawed. While major publications in the LLM detectability field suggested that LLMs were easy to detect with fine-tuned autoencoders, the limitations of their results are easy to overlook. Specifically, they assumed publicly available generative models without fine-tunes or non-trivial prompts. While the importance of these assumptions has been demonstrated, until now, it remained unclear how well such detection could be countered.

Here, we show that an attacker with access to such detectors' reference human texts and output not only evades detection but can fully frustrate the detector training - with a reasonable budget and all its outputs labeled as such. Achieving it required combining common "reinforcement from critic" loss function modification and AdamW optimizer, which led to surprisingly good fine-tuning generalization.
Finally, we warn against the temptation to transpose the conclusions obtained in RNN-driven text GANs to LLMs due to their better representative ability.

These results have critical implications for the detection and prevention of malicious use of generative language models, and we hope they will aid the designers of generative models and detectors.


\end{abstract}

\section{Introduction}

Ever since the introduction of the Transformer modular self-attention architecture \citep{Transformer2017Google}, the training parallelization this architecture allowed led to a continuous scaling of models - be it in the number of parameters, computational resources invested into training them, and training datasets beyond what was previously imaginable \citep{DistillBERT}. Earning them the name of Large Language Models - LLMs, such architectures are ubiquitous in modern NLP ML. Be it GPT family optimized for the text generation \citep{GPTPaper2018,GPT2Paper2019,GPT3Paper,GPT4Paper}, BERT/RoBERTa optimized for gap-filling and classification \citep{BERT2019,RoBERTa2019}, or T5 \citep{T52020} optimized for translation-like generation, LLMs successfully scaled across more than four orders of magnitude in parameter and training dataset size, all while continuously improving in their primary task.

Unexpectedly, in addition to continuously improving in the task they were trained for, LLMs also  underwent a qualitative transition in their capabilities, unlocking unexpected and seemingly unrelated capabilities. Learning basic arithmetic \citep{GPT3Paper}; generating valid Python programs from natural language requirements \citep{LargeLanguageModelsProgramming2021}; predicting recidivism better than specialized models and generating unprompted hate speech \citep{Antropic2022BiggerModelsUnlockPerformanceAndRacism}; - all were achieved without any further model modification, through prompt choice alone.

Perhaps the most impressive achievement of LLMs was their ability to generate natural language so well that some LLMs could not be distinguished from humans in most settings. While GPT-2 was already becoming good at this \citep{Ippolito2020HumansAreEasilyFooled}, it was GPT-3's ability to run popular wellness advice blogs \citep{hackernews_well_being_GPT3_2020} and write newspaper articles about itself \citep{guardian_gpt3_2020} that led to wide-reaching concerns about the detectability of LLM outputs. The situation got only worse with the release into open access of conversationally fine-tuned LLMs \cite{InstructGPT2022OpenAI}, making LLMs accessible and easy to use to the general public, and with the augmentation of LLMs with auxiliary capabilities, such as information retrieval, calculator and image analysis \cite{SeeKeR2022Facebook,LaMDA2022Google,GPT4Paper}.

Such performance is impressive, but it poses a serious threat. Generative models have no inherent moral code, and safety features added by LLM designers to imitate it are easily bypassed. LLMs can trivially be used to run blogs and social media accounts and write articles pushing disinformation, supporting harassment campaigns, or promoting self-harm. 
Undercover social influence and harassment with economical \citep{Luca2016FakeReviewsEconomy}, political \cite{hotez_anti-science_2021}, and military goals \citep{Dawson2019InternetResearchAgencyDesinformation} is a well-developed and highly active industry \citep{bradshaw2021industrializedDesinformation,Bagdasaryan2021SpinningLM}. Everything points towards generative language models being able drive an automation revolution in it \citep{TruthLiesAndAutomation2021,Stiff2022DetectingComputerGeneratedDesinformation}. Even relatively small, widely accessible models that can be run on commodity hardware can be extremely effective in generating misleading news articles \citep{Zellers2019Grover,Kreps2020All},
writing fake reviews \citep{Adelani2020GeneratingSentimentPreservingFakeOnlineReviews}, phishing \citep{Mink2022DeepPhish}, or disrupting of democratic processes
\citep{Weiss2019DeepfakeBotPublicFedWebsite}.

Despite the best intentions of their original authors, generative language models pose a major threat in case of misuse. It is essential to be able to detect them in the wild. 

\section{Background}

\subsection{Generated Text Detection}

Prior research into generative language model detection is extensive and predates LLMs \citep{Hovy2016FakeReviewsHMMs}. Model detection concern was so great for the creators of GPT-2 model that they provided a baseline detection tool for their model \citep{Solaiman2019ReleaseStrategiesOfLanguageModels}. Creators of GROVER - an LLM specializing in news-like text generation - claimed that it also could detect generative models at large \citet{Zellers2019Grover}. Such detection tools were deemed highly necessary, given that even for relatively small LLMs, humans were shown to be inadequate at detecting them \citep{Ippolito2020HumansAreEasilyFooled}. Several high-profile 
benchmarking studies found that fine-tuned classifier LLMs performed best\citep{Solaiman2019ReleaseStrategiesOfLanguageModels,Zellers2019Grover,Ippolito2020HumansAreEasilyFooled,Stiff2022DetectingComputerGeneratedDesinformation,Maronikolakis2021IdentifyingArtificiallyGeneratedHeadlines,Adelani2020GeneratingSentimentPreservingFakeOnlineReviews,Diwan2021FingerprintingFineTunedModelsInTheWild,Fagni2021TweepFake}. Such detectors were trained by feeding a pretrained LLM, often a BERT/RoBERTa, with examples of machine-generated and human-generated texts, hoping that the language representation learned during the pretraining will allow the LLM to learn robust features that are specific to either of classes.

However, most work on generative model detection did not consider the possibility that attackers would fine-tune their generative models to evade detection or prompt them in a non-trivial manner. A flurry of research papers moderating the initial optimistic results followed. A model fine-tuned for unrelated purposes was shown to evade detectors trained on the base model \citep{Adelani2020GeneratingSentimentPreservingFakeOnlineReviews}. Similar results could be achieved by using longer prompts \citep{Bakhtin2019RealOrFakeLearningToDiscriminateMachine}, changing the sampling strategy \citep{Ippolito2020HumansAreEasilyFooled,Stiff2022DetectingComputerGeneratedDesinformation}, or adversarially perturbing characters or words in the prompt \citep{Gagiano2021RobustnessOfGrover,Wolff2020AttackingNeuralTextGeneration}.

Even in the white-box setting - when the generative model is identified, and its outputs are correctly labeled - SotA methods struggle to detect them. \citet{Adelani2020GeneratingSentimentPreservingFakeOnlineReviews} showed that in cases where fine-tuned models were known in advance, a BERT-based detector could be trained to detect them, although with low precision and recall  (~40\% for both). \citet{Fagni2021TweepFake} found that even for Twitter bots based on generative models that do not attempt evasion,  
even fine-tuned detectors struggled to differentiate GPT2-based ones from humans (~70\% accuracy). 
Perhaps more concerning, the memorization capabilities of generative LLMs mean that well-designed prompts can trigger perfect recall of training data - which is human-generated text \citep{GPT2LeaksAF}. A detector would need access to the whole training dataset of the generative model to succeed in that setting, which is an unrealistic assumption, especially in an adversarial detection setting.

\subsection{Detection and Evasion in Text GANs}

A common framework to approach an arms race of detection and detection evasion in ML is Generative Adversarial Networks (GANs) \citep{GoodfellowGANs2014}. They are notorious for excellent capabilities in image generation, with photorealism and style transfer often cited as examples of performance \citep{BigGAN,DayToNightConversionGAN2017}. For natural language generation, GANs have been investigated as ways to improve pretrained models, to further improve models trained through maximum likelihood methods - specifically mitigate the exposure bias. Exposure bias is a tendency of language models to generate a succession of tokens they never encountered in training and find themselves without a statistical model to continue generation, leading to repetitive, degenerate output \citep{ScheduledSamplingModeCollapse2015,GenerativeModelsDegeneration2020}.

Unfortunately, despite a wealth of proposed architectures, none of the text-generating GANs imitating image-generating ones offered any improvement over MLE \citep{Caccia2020LanguageGANsFallShort}. While several modern GANs have been proposed to mitigate that, such as ScratchGAN \citep{ScratchGAN2019} or ColdGAN \citep{Lai2020ColdGAN}, they depart significantly from the traditional adversarial setting and require a collaboration between the generator and discriminator, which is an unrealistic assumption for in-the-wild detection.

More importantly, all text-generating GANs proposed until now that are suitable for the adversarial setting are based on RNNs. Prior research suggests that proposed architectures stop working upon a transition to self-attention-based LLMs \citet{KevinPaper2021}.

To illustrate this issue, we start with a Diversity-Promoting GAN (DPGAN) \citep{Xu2018DPGAN}, that combines rewards for words and the whole sentence and corresponds to the scenario where the attacker would have access to both overall human/machine score and single word influence on the score. This is a realistic scenario for an attacker seeking to bypass a standard, widely accessible generative models detector, such as the HuggingFace OpenAI detector, a common tool that gained popularity thanks to its public accessibility, or GLTR - an early generative text detection tool \citep{Gehrmann2019GLTR}.

\subsection{Training Stochastic Parrots: Reinforcement from Critic}

Adjusting LLMs to improve a specific aspect of their behavior is a common practice and is often done through further model training - \textit{fine-tuning}. Reinforcement Learning from Human Feedback is an example of such fine-tuning that uses an external critic model trained from human feedback \citep{InstructGPT2022OpenAI}, but the general approach is significantly older \citep{Ziegler2019FineTuningLanguageModelsFromHumanPreferences}.

Reinforcement from a critic model has been particularly prominent in the generative model normativity alignment field \citep{Sheng2020NormativeGenerationReview}. LLMs are trained on large datasets of texts collected on the internet and are representative of those texts. Because of that, they are prone to unexpectedly biased and toxic text generation \citep{Bolukbas2016ManIsToComputerProgrammer,Antropic2022BiggerModelsUnlockPerformanceAndRacism}. A term of \textit{Stochastic Parrots} has emerged to designate this tendency \citep{StochasticParrotsGebru2021}, and one of the approaches to counter it is normative fine-tuning. The principle of normative fine-tuning is to use a critic LLM trained to predict when non-normative text is being generated and fine-tune the generator LLM from the feedback of the critic LLM, using rewards as a custom loss \citep{Solaiman2021PALMSdatasetFineTuning,ReducingNonNormativeTextGeneration2020,Ziegler2019FineTuningLanguageModelsFromHumanPreferences}.

Here, we provide a setting that combines the reinforcement from a critic model with a more modern AdamW optimizer\cite{Ma2019AdamW}, rather than the traditional Adam one \citep{Kingma2015Adam} to lead to a robustly generalizing fine-tune. We start by performing a normativity fine-tune using a sentiment classifier - an imperfect, although valid approach \citep{ReducingNonNormativeTextGeneration2020}. After confirming our approach, we swap the normativity critic model for a generative model detector.

\section{Contributions}

\begin{itemize}
  \item We show that generative LLM detection with a discriminator LLM is impossible if the attacker has access to the reference "human" dataset used to train the discriminator LLM. Simply fine-tuning the dataset and using prompts from it leads to a complete failure of the discriminator to learn the difference between machine and human-generated texts, even in a setting where all LLM outputs are correctly labeled.
  \item We show that reinforcement from critic generalizes significantly better than previously described when paired with the AdamW optimizer rather than the commonly used Adam one and allows well-generalizing model fine-tunes from limited data, matching prior SotA in normativity fine-tuning.
  \item We demonstrate a critical weakness on a previously proposed text-generating GAN architecture - DPGAN, and show the connection of this weakness to the difference in representative power of LLMs and RNNs used in text GANs compatible with the in-the-wild detection setting.
\end{itemize}

\section{Methodology}

Our code is based on a pre-existing Pytorch implementation of common text-generating GANs, including DPGAN, available from \url{https://github.com/williamSYSU/TextGAN-PyTorch}.

We chose GPT-2 small (117M parameters) as a generator due to its wide availability and extensive research record concerning its usage. In addition to similar considerations for the choice of BERT base (110M parameters) as a discriminator, fine-tuned BERT is close to SotA among methods for generative model detection in the wild \citep{Diwan2021FingerprintingFineTunedModelsInTheWild,Fagni2021TweepFake}. Overall, the configuration represents a likely attacker/defender language model configuration when both are limited to commodity hardware. 

We rely on two datasets provided by the base text-GAN generating library, the 10 000 entries MS COCO scene description dataset and the 280 000 entries 2017 EMNLP news sample corpus. The first five tokens are used as prompts whenever applicable, and we use an 80/20\% train/validation set. No hyperparameter search was performed, with default parameters and parameters from prior literature being used. Adam \citep{Kingma2015Adam} and AdamW optimizers \citep{Loshchilov2019AdamW,Ma2019AdamW} were used, according to subsections of \ref{ResultsDiscuss}. Per prior research on the importance of model training re-starts for methodology evaluation \citep{Bosquet2018GANsEqualUpToRestart}, we report all model training runs and evaluate the model performance on the best-out-of-five.

A more detailed overview of the methodology is available in Appendix \ref{DetailedMethodsAppendix}. The code and detailed instructions to reproduce all experiments can be found at \url{https://github.com/8a3539f168fd077097ea473cc8a9c093/gpt_bert_gan}.

\section{Results and Discussion}
\label{ResultsDiscuss}

\subsection{GPT-2 Collapses Rapidly Due to DP-GAN's Discriminator Misspecified Reward}
\label{subsec:5.1}

DPGAN attributes to generated samples a sentence-level reward and a word-level reward for each of the generated tokens to favor/discourage the generation of tokens at some position. By replacing the initial generator with a GPT-2 implemented based on prior code (more details in Appendix \ref{DetailedMethodsAppendix}), we train GPT-2 from scratch with the MS COCO dataset, while keeping the same discriminator. We expect full memorization from GPT-2 trained from scratch and expect the discriminator to fail and provide overall rewards favoring rare words. Surprisingly, we noticed consistent output degeneration after only two epochs, ie. GPT-2 only generating 'a' tokens. This contradicts the original paper suggesting that DPGAN was specifically designed to avoid repetitive common tokens in its output \citep{Xu2018DPGAN}. As we can see in table \ref{output_degeneration_Table_prob}, the generator quickly starts to prioritize the generation of the token 'a' at position two, even though otherwise rewards indeed favor more rare words, as authors originally suggested. While almost all samples in MS COCO start with an 'a', it should have a probability of almost one only at the first position and certainly not the second.
% Probability token at position 2

\begin{table}[!h]
	\centering
	\begin{adjustbox}{width=0.8\columnwidth,center}
	\begin{tabular}{ |c||c||c||c| }
 		\hline
		\multicolumn{4}{|c|}{Probability of token at position 2} \\ 
		\hline
		epoch 0 & epoch 1 & epoch 2 & epoch 3 \\
		\hline
		 woman: 0.31 & woman: 0.46 & a: 0.66 & a: 0.99 \\
		 view: 0.14 & is: 0.19 & is: 0.31 & is: 0.0001 \\
		 corner: 0.05 & kitchen: 0.05 & woman: 0.009 & woman: 0.000002 \\
		 kitchen: 0.04 & white: 0.04 & kitchen: 0.009 & kitchen: 0.000001 \\ 
		 bathroom: 0.04 & cat: 0.02 & white: 0.003 & white: 0.00000008 \\
		 \hline
		%Different stuff & 99 & 12 \\
	\end{tabular}
	\end{adjustbox}
        \vspace*{0.25 cm}
 	\caption{Tokens with the highest probability of being generated by GPT-2 at position two during adversarial training}
  \label{output_degeneration_Table_prob}
\end{table}

After noticing the repetition of the token 'a', we inspected the rewards the discriminator gives to generated samples. We noticed that token 'a' was favored only in the first position, while the rest of the text favored less frequent tokens (see table \ref{output_degeneration_Table_rewards}). In particular, the discriminator outputs positive and negative rewards, but these rewards seem to be too high for 'a', which likely overwhelms the word-based reward of the DPGAN discriminator.

% word rewards
\begin{table}[!h]
	
	\centering
	\begin{adjustbox}{width=0.7\columnwidth,center}
	\begin{tabular}{ |c|c|c|c|c|c|c|c|c|c| }
		\hline
		\multicolumn{8}{|c|}{epoch 0} \\ 
		\hline
		 BOS & a & woman & is & walking & across & the & street \\
		\hline
		33 & 868 & 0.4 & 0.01 & 1.2 & 0.4 & 0.4 & 1.8 \\
		\hline
		%Different stuff & 99 & 12 \\
	\end{tabular}
	\end{adjustbox}
	\newline
	\vspace*{0.25 cm}
	\newline
	\begin{adjustbox}{width=0.7\columnwidth,center}
	\begin{tabular}{ |c|c|c|c|c|c|c|c|c|c| }
		\hline
		\multicolumn{9}{|c|}{epoch 1} \\ 
		\hline
		BOS & a & large & gray & airplane & flying & through & the & sky \\
		\hline
		33 & 869 & 5.6 & 6.6 & 11 & 10 & 1.7 & 2.8 & 7.5 \\
		\hline
		%Different stuff & 99 & 12 \\
	\end{tabular}
	\end{adjustbox}

	\vspace*{0.25 cm}

	\begin{adjustbox}{width=0.55\columnwidth,center}
	\begin{tabular}{ |c|c|c|c|c|c|c|c|c|c| }
		\hline
		\multicolumn{8}{|c|}{epoch 2} \\ 
		\hline
		BOS & a & a & a & a & a & a & a \\
		\hline
		 33 & 870 & 0.6 & 0.07 & 0.03 & 0.04 & 0.05 & 0.07\\
		\hline
		%Different stuff & 99 & 12 \\
	\end{tabular}
	\end{adjustbox}
        \vspace*{0.25 cm}
        \caption{Rewards given to the first generated tokens by the discriminator}
        \label{output_degeneration_Table_rewards}
\end{table}

By doing the same training with different variations of the MS COCO dataset (see appendix \ref{dpgan_debug}), we arrived at the conclusion that, in our setup, the discriminator favored the generation of tokens that are frequently present in the dataset due to unexpected behavior of the discriminator.

This, however, brought forwards the question of why the problem has not been discovered and reported in the original paper. Given that the problem arose when we replaced an LSTM generator with a larger and more powerful GPT-2, we investigated whether the size/representative power of GPT-2 relative to the training dataset was an issue. Experiments with different sizes of GPT-2 and a larger dataset (details about different sizes can be found in appendix \ref{lighter_gpt2}). Even though we observed a small drop in quality(repetition of tokens), we didn't observe the same degeneration as with the larger model and no evolution of the model at all in the case of a larger dataset (as can be seen in Supplementary Fig. \ref{BLEU_evolution}). We believe that this confirms our hypothesis and is a more general problem to which all RNN-based text GANs would be susceptible when converted to self-attention-based LLMs.

\subsection{Making Sure a GAN Setting Allows the Generator to Train}
\label{subsec:5.2}

Given that we did not see any model evolution and a discriminator failure in DPGAN, we decided to completely change the architecture and first validate it with a fixed discriminator to show that a critic-guided fine-tuning of GPT-2 worked as expected. Given that a common setting for such critic-guided fine-tuning is normativity alignment, we used a previously proposed approach to it - using a sentiment analysis classifier. The rationale behind this approach is that non-normative text would be associated with a negative reaction to it, leading the sentiment classifier to detect and suppress it \citep{ReducingNonNormativeTextGeneration2020}. This would allow us to not only validate our approach  for evasion fine-tuning but also provide results for the normativity alignment field. The details about how GPT-2 is fine-tuned can be found in Appendix \ref{negativity_reduction_training}.

In the experiments described in this part, the "BERT model" is a pretrained model fine-tuned for sentiment analysis. In this section, we will call \textit{nice GPT-2} a base GPT-2 model fine-tuned using the sentiment classification BERT model, with the generation being prompted by the first five tokens in the MS COCO dataset. We  call \textit{base GPT-2} a default pretrained GPT-2-small model without any  fine-tuning.
 
We experimented with different BERTs trained for sentiment analysis. The difference between them was in the labels they gave to generated samples. We found that having a BERT model that could attribute a third "neutral" label (mapped to a loss reward of 0.5 in our case) to the generated samples was helpful for the training. Most likely, this helped because most sentences in the MS COCO dataset were neutral, given that MS COCO entries are generally neutral image descriptions.

\begin{figure}[h!]
\centering
\includegraphics[width=0.65\textwidth]{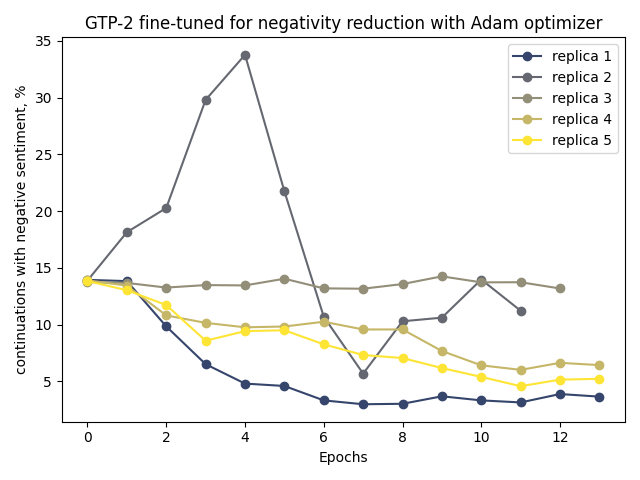}
\caption{Number of samples classified as negative by BERT during training at each epoch with 10,000 prefixes for three different runs}
\label{training_sentiment_Adam}
\end{figure}

We used Adam optimizer with $betas = (0.9, 0.999)$, and $epsilon = 1 \cdot 10^{-08}$. For the loss function, we used MAE (mean absolute error). We already observed some meaningful but unstable results while training with the MS COCO dataset. Out of 5 runs, we observed one where the negativity didn't change, three that had about 70\% less negativity, and one that fluctuated between more than double negativity and 50\% less negativity(see Fig. \ref{training_sentiment_Adam}). 

\begin{figure}[h!]
\centering
\includegraphics[width=0.65\textwidth]{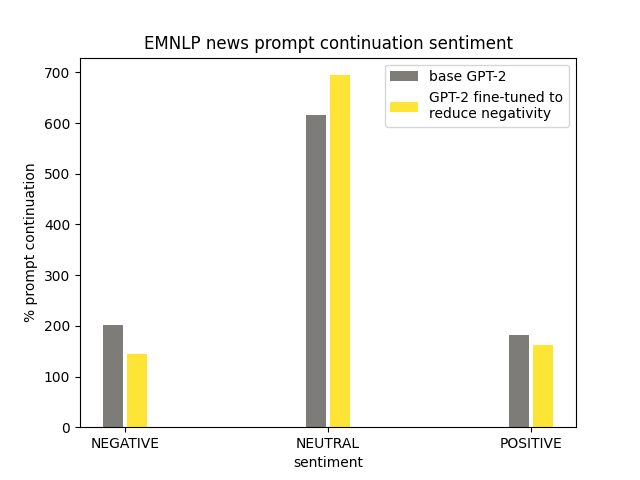}
\caption{Comparison of the sentiment of text generated by base GPT-2 and finetuned version on a reduced version of EMNLP dataset}
\label{emnlp_validation_Adam}
\end{figure}

We then tried to see if our result would generalize to another dataset, so we compared pre-trained GPT-2 with our nice GPT-2 on the EMNLP news dataset. We observed that the result generalized to about 35\% less negativity with GPT-2 nice compared to base GPT-2 (see Fig. \ref{emnlp_validation_Adam}), which is less than the 70\% less negativity observed for the training set(MS COCO). This is comparable to the prior SotA from \citet{ReducingNonNormativeTextGeneration2020}. We also provided samples generated by base GPT-2 and nice GPT-2 in appendix \ref{samples_gpt2_sentiment}. We believe that GPT-2 sometimes learns during training to avoid generating types of sentences considered negative by the BERT, and this procedure generalizes to a validation dataset.

We noticed that, on average, the base GPT-2 generates more negative samples than positive ones, and some of them are very negative (we provide a selection of them in appendix \ref{samples_gpt2_sentiment}). This considerable bias should be accounted for when using GPT-2. It is now well known that some prefixes lead to extreme toxicity by the GPT family \citep{gehmanEtal2020Realtoxicityprompts}, but it seems that sometimes prefixes that we would not suspect also lead to very toxic generated text. This is not surprising and has been previously reported by \citet{RedTeamingLanguageModels2022}, although the problem is known to be significantly worse for larger models of the GPT family (see \citet{Antropic2022BiggerModelsUnlockPerformanceAndRacism}).

\subsection{Ensuring Generalizable Generator Training}
\label{subsec:5.3}

The problem with the prior approach was that in about 2/5 cases, the training did not lead to improvement at all. We observed on some runs an increase in negativity up to double or no changes in negativity during training. Also, the negativity reduction on the validation set(EMNLP news) was 35\% compared to the 70\% reduction for the training set. The instability of large language model training is well-known, especially on smaller datasets, such as MS COCO (multiple restarts are often used to obtain state-of-the-art models). However, we were unsatisfied with this performance. We tried to use tricks known to stabilize training and increase the generalization capabilities of the model, namely switching to use \emph{AdamW} for the optimizer, learning rate scheduling(with warm-up), gradient clipping, and weight decay, all of which were used for GPT-2 pretraining. We also tried to use SGD, which is known to generate results that generalize better and allow model "grokking" \citep{Power2021Grokking}. Unfortunately, SGD is known to fare poorly on language models and failed outright.

\begin{figure}[h!]
\centering
\includegraphics[width=0.65\textwidth]{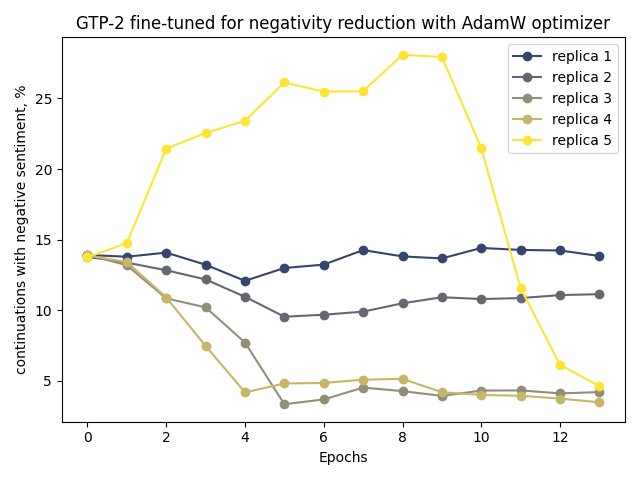}
\caption{Number of samples classified as negative by BERT during training at each epoch with 10,000 prompts for five different runs}
\label{sentiment_training_adamW}
\end{figure}

We ran our model with 14 epochs(we stopped early when we saw no changes for five epochs). We tested learning rate scheduling with linear scheduling and a starting learning rate at 5e-5(the learning rate is reduced linearly ten times per epoch until reaching 0 at epoch 20)). The reason we use a learning rate scheduling is to improve stability because it allows the model to reach a more robust and deeper local minimum. For the other parameters(in particular for the parameters of AdamW, we looked at default parameters from:\url{https://github.com/huggingface/transformers/blob/main/src/transformers/training_args.py}.

\begin{figure}[h!]
\centering
\includegraphics[width=0.65\textwidth]{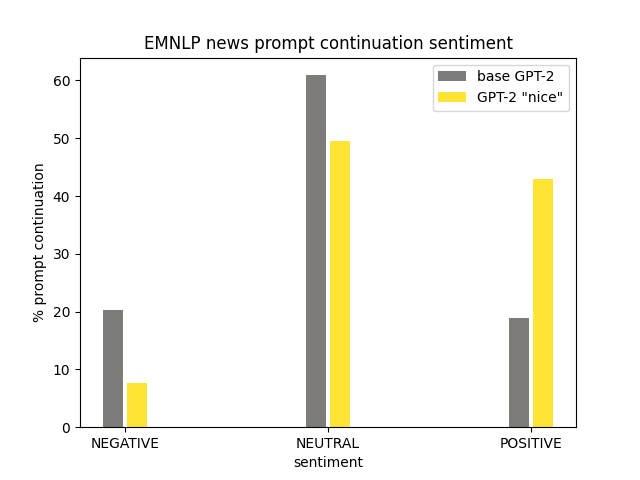}
\caption{Comparison of the sentiment of text generated by base GPT-2 and nice GPT-2 with a generalizing improvement in positivity tested on the EMNLP news dataset}
\label{validation_adamW}
\end{figure}

Using the above tricks, we observe only marginal improvement in stability. We still observed the same rate of runs that did not lead to improvements. However, we saw better convergence towards about 5\% less negativity for the training dataset(for runs that did reduce the negativity). We also observed better generalization for the validation set with 60\% reduced negativity. In particular, one of the training produced a particularly good model that generalized well with the validation EMLP dataset to an approximate 60\% decrease in negativity and a 60\% increase in positivity(see Fig. \ref{validation_adamW}). We thus think that the model can get similar levels of reduction of negativity for the validation set when the GPT-2 generator is able to train. However, even with different optimization tricks, we observed the same rate of about 2/5 runs that did not lead to less negativity.

Relevant literature suggests that transformer NLP models training and fine-tuning are hard tasks that require extensive hyper-parameter optimization and additional tricks to stabilize the training(see eg. \citet{Liu2020UnderstandingTransformerTrainingDifficulty}).

\subsection{Running GAN: Hide-and-Seek Between GTP-2 and BERT}
\label{subsec:5.4}

The goal of these experiments is to see if it is possible to train a BERT model to detect text generated by GPT-2 and use it to then train GPT-2 to escape training, aka perform GAN iterations.

\begin{figure*}[ht]
\centering
\includegraphics[width=0.85\textwidth]{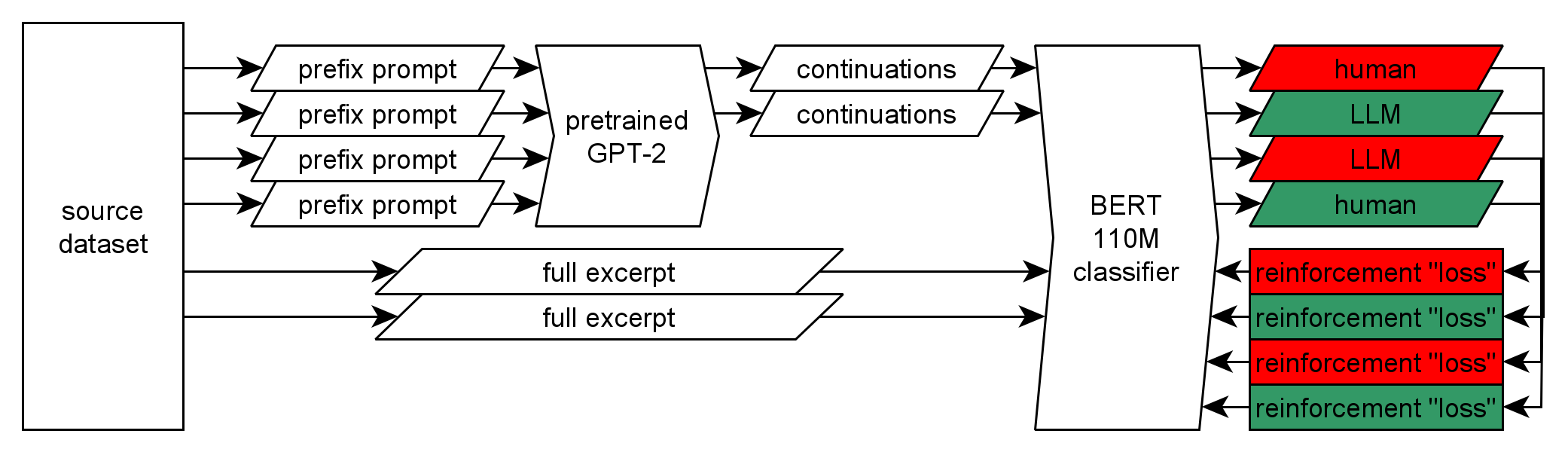}
\caption{BERT training phase of the GAN for fake detection}
\label{bert_detection_fake}
\end{figure*}

The architecture we designed is functionally divided into two parts. First, BERT fine-tuning, which we will call \emph{discriminator training} (see Fig \ref{bert_detection_fake}) and  generator training with scores from BERT, which we will call \emph{generator training}(see Supplementary Fig. \ref{gpt2_fake}).

The BERT fine-tuning (discriminator training) consists of fine-tuning a pre-trained BERT classifier to detect fake samples. We feed BERT with samples generated by a base pre-trained GPT-2 that uses prefixes from a dataset to generate samples. We give the label 0 to fake samples generated by GPT-2 and label 1 for the true samples coming from the same dataset as the prefixes that GPT-2 uses. Then BERT outputs a probability of the output being fake or true given an input sample. We then compute a loss based on the labels and perform back-propagation. The process is repeated for 1-3 epochs, where each epoch corresponds to the full dataset traversal. 

The second part of the training(generator training) consists of training GPT-2 with scores exactly the same way as we did for sentiment analysis(cf. part \ref{subsec:5.2}). The details about how it differs from the previous part can be found in appendix \ref{generator_training_fake}.

The GAN architecture we designed starts with an optional fine-tuning of the GPT-2 generator then we repeat parts 1 and 2 during the training.

\subsubsection{Training with Fine-Tuned Generator}

We first tried to fine-tune GPT-2 for the MS COCO dataset before the two repeated phases mentioned above, then built a dataset from the true MS COCO dataset and fake MS COCO generated by fine-tuned GPT-2. This dataset is used to fine-tune BERT to detect fake samples. Finally, use BERT fine-tuned for fake detection to train GPT-2 as we did in the previous experiment with sentiment analysis(the goal being to build a GAN from that).

\begin{figure}[h!]
\centering
\includegraphics[width=0.65\textwidth]{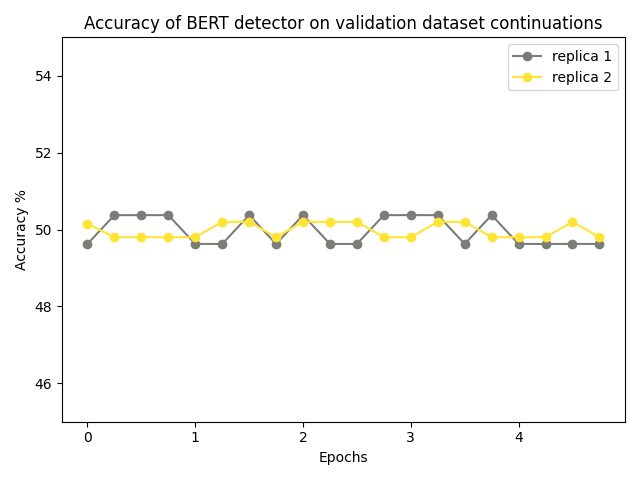}
\caption{BERT accuracy on the validation set when GPT-2 is fine-tuned}
\label{bert_acc}
\end{figure}

The result was that BERT could not distinguish samples generated by GPT-2 from true samples(see Fig. \ref{bert_acc}). Also, we discovered here that BERT achieved almost 100\% accuracy when the samples generated by GPT-2 have a different length than those in the true dataset (more details about this training setup can be found in appendix \ref{fake_detection_training_fine_tuned}).

\subsubsection{Training Without a Fine-Tuned Generator}

Since we found that BERT could not train well when the difference between fake and true datasets is small(one or two tokens) or too big (samples generated from GPT-2 have ten more words than those of the dataset), we decided to perform the same procedure, but without fine-tuning GPT-2, and with correction of the lengths of the generated samples(so that all samples true/fake have same lengths). In that scenario, BERT had a high accuracy of about 90\%. However, GPT-2 was not able to learn to evade BERT fine-tuned for fake detection (see appendix \ref{fake_detection_training_no_fine_tune} for more details about this part).

Our hypothesis as to causes is two-fold. First, we saw in part 5.2 that training GPT-2 with scores, as we did, is unstable (this could probably be improved). Also, we inspected the scores that BERT gave to the generated samples(see appendix \ref{score_samples} for examples of samples and scores), and we didn't observe a general pattern in the samples that received a good score. GPT-2 is, therefore, probably not able to find types of samples to avoid or to generate more of to fool BERT (as it maybe did in part \ref{subsec:5.2} because the scores were more meaningful). 
BERT and GPT-2 might need more data to generalize for fake detection. In particular, transformers require a lot of data to be trained (see \citet{PopelTransformerTips2018}), and GPT-2 might need to be further fine-tuned by using prompts derived from the training dataset.

\subsubsection{Training with a Partially Fine-Tuned Generator}
For the reasons above, we combined EMNLP news and MS COCO as a single dataset, and we fine-tuned the generator using MLE training with 30\% of that dataset before adversarial training. During the discriminator training, BERT achieved 88\% accuracy in fake detection on the validation dataset. The result for the adversarial training in terms of the number of samples generated by GPT-2 classified as fake or true is shown in Supplementary Fig. \ref{gpt2_fake_scores_emnlp} in Appendix. Although GPT-2 is able to learn to evade fake detection, it degenerates during training (as can be seen with the samples in appendix \ref{samples_emnlp_coco}).

\section{Conclusion}

Our overarching result is the demonstration of a highly potent attack against a common type of generative LLM detectors, which are considered SotA and are highly used in the wake of ChatGPT public release in late 2022.

Specifically, we show that an attacker with access to the human reference texts used to train the detector and access to the detector rating of generative model outputs is able not only to evade the model detection but even in the white-box setting, where all the generative mode outputs are correctly labeled, they can fully stop the detector training against it. While the detection of generative LLMs with other LLMs has been previously shown not to work well, we go further, showing that in the setting where common datasets are used to train the discriminator, a minimally competent attacker can fully defeat the detector. 

On the way to this result, we showed that reinforcement from a critic model could be used to fine-tune a generative model on a relatively small dataset, as long as AdamW rather than the common Adam optimizer was used, with multiple restarts yielding the best results.

Finally, we showed that existing literature on text-generating GANs, with both generator and discriminators built on RNNs, cannot be trusted to translate to LLMs as-is. While we demonstrate a failure in the DP-GAN architecture specifically, the reasons we believe led that failure to have gone undetected are the fundamental difference in the representative power of GANs and RNNs. As such, careful re-validation is needed. 

Overall, we believe that our results strongly argue against the continued use of LLMs fine-tuned for classification to detect generative LLMs in the wild. We hope that this prompts research into novel methods of LLM detection, even if it is just fingerprinting of texts from major generative LLM access providers.

We hope that our auxiliary results will also be of interest to the NLP ML community, helping fine-tune generative models more efficiently and will prompt the re-examination of results obtained on RNN-based text GAN before their extrapolation to LLMs.

\section*{Limitations}

All the results presented here are primarily experimental. We do not provide theoretical guarantees, and it is unclear how results obtained on GPT-2 small and BERT will scale up to larger models, given their tendency to unlock new capabilities with the increase in size \citep{Antropic2022BiggerModelsUnlockPerformanceAndRacism}. Similarly, while prior research suggests that results obtained on those two specific architectures generalize to other Transformer-based architectures, it remains an open question. GPT-2 small and BERT are architectures of comparable sizes (117M parameters vs. 110M parameters). An adversarial training involving models of significantly different representative power would likely result in different dynamics. A notable example of such a setting is the case where the discrimination model is substantially smaller than the generative one - which is a likely setting, given the attacker can invest considerable computational resources, whereas due to the sheer volume of content generated and consumed by humans online, a defensive discriminator will likely need to run on end-user devices.

While large language models fine-tuned for detection - such as the ones we used here - have been criticized as stylometry unsuited for generative models detection \citep{Bhat2020SemanticsNotSyntax,Schuster2020StylometryLimitations}, there has so far been no alternative detection method shown to perform better. Notably, methods relying on the factual structure of the text have shown similar vulnerability to prompt selection and fine-tuning \citep{Zhong2020DeepfakesFActualStructure}, even before the arrival of factual database augmented generative models \citep{Shu2021FactEnhancedNewsGeneration,SeeKeR2022Facebook}.

Similarly, evasion detection through prompt selection - that we use to some extent - has been almost entirely neglected until now. However, results from the generative model normativity and privacy fields suggest that well-chosen prompts can lead to highly unexpected and uncharacteristic texts \citep{RedTeamingLanguageModels2022,GPT2LeaksAF}, likely leading to a degradation of detection capabilities presented here as a previous SotA. 

If anything, both of those avenues for evasion detection reinforce our conclusion that a competent attacker can easily evade detection even with relatively small models.

\section*{Ethics Statement}

We essentially provide a blueprint for a competent attacker to create a generative model that would effectively evade detection by most if not all, means available today.

While the first version of this paper was prepared in mid-2022, following the release of ChatGPT, we delayed its submission for publication by four months following the public release of ChatGPT to leave time for model developers to improve detection mechanisms. We similarly have contacted some of the entities operating common generative LLM detection endpoints to report our findings, to no avail. 

Given that prior knowledge of reference "human" texts used to train the detector, LLM is a critical component of full evasion described here, and given that it is currently unlikely for most deployed detectors using classification fine-tuned LLMs, we decided in favor of releasing our results publicly.

Another factor that contributed to our decision to release these results is an increasing reliance on fundamentally flawed detection tools to make prejudicial decisions. We observed LLM-based detectors for generative LLMs used in the educational setting, meaning that their high false-positive rate led to students being unjustly accused and disciplined, potentially impacting their long-term perspectives. 

We hope the results presented here will lead to a more rigorous study of alternative ways to detect LLMs, starting with the generated text fingerprinting by major generative LLM providers.

\section*{Acknowledgements}

We would like to thank the armasuisse - Cyber-Defence (CYD) Campus for the Distinguished Post Doctoral Fellowship supporting AK, as well as Fabien Salvi (EPFL) for the technical support regarding the computational infrastructure organization, and France Faille (EPFL) for the administrative support.

\bibliographystyle{splncs04}
\bibliography{custom.bib}

\newpage

\appendix

\section{Detailed Methodology}
\label{DetailedMethodsAppendix}

\subsection{GAN configuration}

We based our code on the Pytorch implementation of a set of text-generating GANs by \url{https://github.com/williamSYSU/TextGAN-PyTorch}, including the DPGAN \citep{Xu2018DPGAN}.

\subsection{Language Models}

We chose GPT-2 model as a generator due to its wide availability and extensive record of usage in natural text generation in different contexts. Besides, GPT architecture scaling across 4 orders of magnitude in parameters with minimal modifications, we can hope for a generalization of our results to larger models. For GPT-2 implementation, we used \url{https://github.com/graykode/gpt-2-Pytorch} as a basis. Unless otherwise specified, we used the GPT-2 small architecture (117M parameters). For part \ref{subsec:5.2} (and also the following parts), we used the weights for GPT-2 from the default HuggingFace GPT-2 repository (\url{https://huggingface.co/gpt2}), but rather than using the \verb|transformers| library provided by HuggingFace, loaded the weights into our implementation of GPT-2 architecture as in \ref{subsec:5.1}. GPT-2 was sampled using a top-40 sampling strategy, starting with the first five tokens in a sentence from a dataset used.

We chose BERT as a base for our defensive discriminator due to its wide availability, extensive usage in creating classifiers for natural languages,  and excellent performance in generative models detection in the past \citep{Diwan2021FingerprintingFineTunedModelsInTheWild,Fagni2021TweepFake}. Overall, the configuration represents a likely attacker/defender language model configuration for both being limited by commodity hardware

\subsection{Datasets}

For fine-tuning and experiments, we used subsets of \emph{MS COCO} and \emph{2017 EMNLP news excerpts  datasets}, provided by the text-generating GAN repository \url{https://github.com/williamSYSU/TextGAN-PyTorch}. In addition to that, as described in appendix \ref{dpgan_debug}, we removed the first token from sentences in COCO and rotated tokens to remove the first tokens

\subsection{Optimizers and Training Parameters}

For all experiments, we used batch sizes varying from 8 to 32 (with accumulation) depending on whether the GPU could support it for training at 14 epochs prior to section \ref{subsec:5.4} and a 1 to 5 epoch after that. Prior to section \ref{subsec:5.3}, Adam optimizer \citep{Kingma2015Adam} was used with default parameters of its PyTorch implementation, with section \ref{subsec:5.3} using AdamW optimizer \citep{Ma2019AdamW} with default parameters of the PyTorch implementation, and a linear learning rate scheduling from 1e-5 to 1e-6 over 10 epochs.

\subsection{Reproducibility}

All the experiments were run on a workstation equipped with Intel Core i9-9900K (8 cores/16 threads CPU), 64 Gb of RAM clocked at 2666 MHz, 2 TB NVME M.2 SSD, and an RTX 3080 graphics cards (10GM VRAM; 29.77 TFLOPS@F16/F32), running an Ubuntu 20.04 LTS distribution. The evaluations were performed within a Docker container, Docker Community Edition, version 20.10.12. The code used Miniconda version 4.12.0, Python 3.9 with CUDA version 11.3.1; PyTorch version 1.10.2, and HuggingFace-hub 0.5. The code and instructions to reproduce all experiments can be found at \url{https://github.com/8a3539f168fd077097ea473cc8a9c093/gpt_bert_gan}.

% \section{Example Appendix}
% \label{sec:appendix}
\section{DP-GAN with GPT-2 Debugging}
\subsection{variation of MS COCO dataset}
\label{dpgan_debug}
After observing that rewards for 'a' were very high, we thought that it was perhaps the fact that almost all sentences start with the 'a' token in the MS COCO dataset that caused the problem. We tested what would happen if the 'a' was moved to the end of the sample of the dataset to analyze the impact of the position of this token (maybe having the same token at the first position very frequently on the dataset could lead to dysfunction while training). However, we found the same result as above. We also tried to remove the 'a's entirely (only the ones at the beginning of the samples from MS COCO). We observed the same outcome as before, ie. a very frequent token in the dataset was being repeated after a few epochs (not 'a' this time but other tokens that were also very frequent in the dataset, such as 'the' or 'woman').
\subsection{lighter GPT-2 and bigger dataset}
\label{lighter_gpt2}
We tried to use 12 layers, 6 layers, and 3 layers(and also 12, 6, and 4 heads, respectively). The total amount of parameters for each configuration is (approximately) therefore 117M, 23M, and 7M, respectively. We have noticed a different behavior with 3 layers compared to 12 layers: while the repetitions of some tokens still happen after a few epochs with 3 layers, we observed no repetitions of a single token compared to the model with 12 layers. We thus think that the size of the model is an important factor in the problems that we observed. 

\begin{figure}[h!]
\centering
\includegraphics[width=0.65\textwidth]{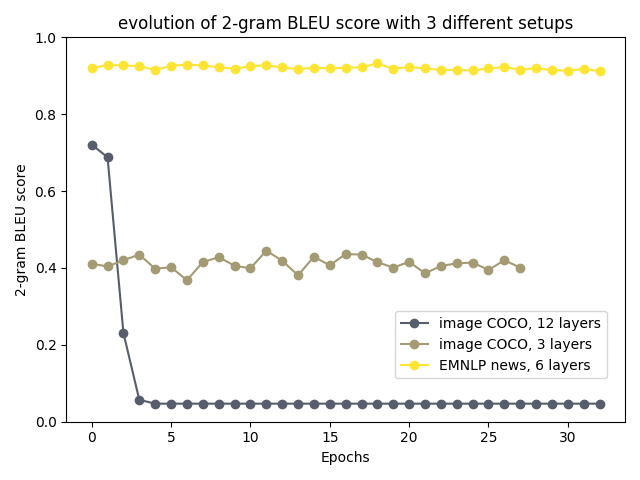}
\caption{evolution of 2-gram BLEU score with different datasets and sizes of GPT-2}
\label{BLEU_evolution}
\end{figure}

We finally tried to see how our model would behave with a bigger dataset. We chose to use the EMNLP news dataset, which is 15 times bigger than MS COCO. The reason to test on another dataset was not only the size of the dataset but also the fact that EMNLP news is more diverse (MS COCO samples are all short descriptions and begin with the 'a' token). We observed an interesting result here. With GPT-2 with only 6 layers, 30 epochs of adversarial training, and the EMNLP dataset as a training dataset: the BLEU score stays the same(see Fig. \ref{BLEU_evolution}). However, we notice no improvement either. One big difference here is that, while GPT-2 (even with 3 layers) is able to memorize(ie. reproduce perfectly samples from the training dataset given a prefix from the dataset), the MS COCO dataset in not more than 30 epochs, GPT-2 with 6 or 3 layers is not able to memorize completely EMNLP news with the same 30 epochs of MLE pre-training and without weights initialization from a pre-trained model.

\section{Examples of Samples Generated by Nice GPT-2 Compared to Base GPT-2}
\label{samples_gpt2_sentiment}
\emph{Warning, some sentences may contain offensive language}
\vspace*{0.5 cm}

Note: These samples were obtained by giving the same prefixes from the EMNLP news dataset to base GPT-2 and nice GPT-2. These tables contain the output of the models given the prefixes. Also, \emph{"improved"} means here that BERT assigned the label "positive" to the generated sample from nice GPT-2, whereas it received the label "negative" when generated by base GPT-2 given the same prefix. 

\newpage
\subsection{improved samples}
\begin{table}[!h]
	\centering
	\caption{samples that were improved when generating them using nice GPT-2 compared to base GPT-2}
	\small
	\begin{tabular}{*{2}{p{.49\linewidth}}}
		\toprule
		\textbf{base GPT-2} &  \textbf{nice GPT-2} \\\midrule
		\emph{"sometimes i think about leaving sex work to my children. I think about having them and my children and being able to do whatever I want to them. And that\'s what I think about when I\'m working with them. I want them to be safe and happy"} & \emph{"sometimes i think about leaving sex work at home in the morning, but i have found it's the right way to deal with this. When i first started i had no idea what to expect from the sex work i had been working on for the past few months. I was"}\\\midrule
		\emph{"I was then stuck in the house with my brother and my sister. It was my first time in the house. My sister told me that my brother's name was Luli and he was going to kill me. My sister did not want to hear this because she knew I would never"} & 
		\emph{"I was then stuck in the house, in the room of a man whose name was Burt and who was the son of a man called J."}\\\midrule
		\emph{"we would open our main presents after lunch, and I'd go back to my room to pick up my own thing. I don't know if I would have done it that way, but I can't think of anything worse than a big pile of things of my own."} & \emph{"we would open our main presents after lunch. The menu is a simple yet effective combination of delicious sandwiches, salads, and even delicious sandwiches. The menu is filled with fresh, locally sourced ingredients, and an extensive menu of seasonal, hand-selected selections. Our selection of' "}\\\midrule
		\emph{"an estimated 80 million people across 20 states and the District of Columbia had been affected by Hurricane Sandy, and as the storm continued to hit the Midwest, the region was especially affected."We\'re seeing more and more of this," said Robert L. Stryker,"} & 
		\emph{"an estimated 80 million people across 20 states will have access to the internet within the next five years. The National Press Club is excited about this announcement that the National Press Club is bringing to all of our readers," said Chris Johnson, Senior Director of Public Affairs for the"} \\\midrule
		\emph{" the 32 - year - old reality star. The story of a young woman who was living in an abusive life in a low-rent, low-temperature home in the foothills of California. She had been a woman of good character, and had been at peace"}	 &
		\emph{"the 32 - year - old reality star who has been a fixture on the big screen since he was 19. However, it is his decision to make his last outing as a celebrity, in the 2008 thriller, The Big Sick, which he directed, that has received the most buzz"}\\
		\bottomrule
	\end{tabular}
\end{table}

\newpage

\subsection{not improved samples}

\begin{table}[!h]
	\centering
	\caption{samples that were not improved when generating them using nice GPT-2 compared to base GPT-2}
	\small
	\begin{tabular}{*{2}{p{.49\linewidth}}}
		\toprule
		\textbf{base GPT-2} &  \textbf{nice GPT-2} \\\midrule
		 \emph{"my sources have suggested that so far the majority of the American population has been the product of a combination of a political system that has been largely corrupted by the corruption of the people. The fact that this system has succeeded in creating an ever-growing majority of the population has been the reason that"} & \emph{"my sources have suggested that so far the Syrian Observatory for Human Rights has confirmed it has been involved in fighting in the past. The Syrian Observatory for Human Rights said that a group of people attacked a house in the town of Al-Wabah, near the Turkish border"}\\\midrule
		\emph{"in only one state , utah , was a commoner than the other states . No God existed in this world . He was created from the Father of all that was in Him and is the same as the Father of all that was in Him and . The Father is the Father of all"} & 
		\emph{"in only one state , utah , which is the Hebrew word for "place of refuge." But there is no place of refuge in our God. We are not our own land or our country, and in the land of my God we are born. But the place of refuge which is"}\\\midrule
		\emph{"there was no immediate claim of responsibility for the attack. In May 2017, a group of people in Syria, including members of the YPG, were massacred by the Syrian army while trying to cross the Euphrates River, a major crossing point for refugees. In July, a group"} & \emph{"there was no immediate claim of responsibility for this, but many have been killed or wounded by the enemy. There have been two other casualties who have been killed by the enemy. The first, a man of the order of the Emperor, who had been wounded in a great fire, is still"}\\	
		\bottomrule
	\end{tabular}
\end{table}

\section{Architectures for Sentiment and Fake Detection Training}

\subsection{generator training in negativity reduction}
\label{negativity_reduction_training}

BERT outputs a sentiment, which is a probability of the input sentiment having the label positive, negative, or, optionally, neutral. For each sample, we transform this sentiment into a score between 0 and 1, by applying the following formula $ P(label=neutral) *(1/2) + P(label=positive) = score$ (where the label is the label that BERT outputs given a sample as input). However, due to how GPT-2 works, we transform this score for the whole sample into a score for the tokens of the samples by giving the whole sentence-level score to each generated token. We then compute a loss between the target of having a sample given the label 1 by BERT(perfect score) for each generated token and comparing it with the score given to the sample.
 
\begin{figure}[h!]
\centering
\includegraphics[width=0.85\textwidth]{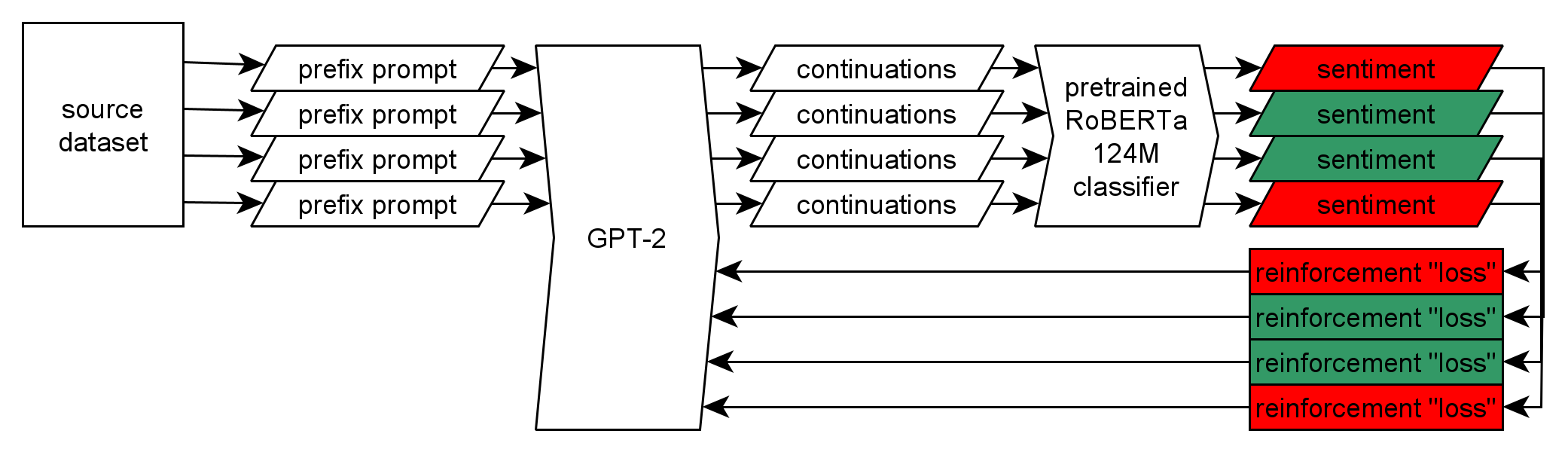}
\caption{GPT-2 with BERT trained for sentiment analysis training}
\label{BERT_sentiment_gan}
\end{figure}

\subsection{generator training in fake detection}
\label{generator_training_fake}
The GPT-2 generator creates samples using prefixes from the dataset, this sample is then fed to BERT. BERT outputs a probability of the sample being fake or true, we then apply a softmax function to the probability output by BERT to produce a value between 0 and 1. This value is what we call the score here(note the difference with the score that we computed in \ref{subsec:5.2}. We finally attribute this score for each token generated by GPT-2 and compute a loss between the score and the target score consisting of only 1s(perfect score) that we use to train GPT-2.

\begin{figure}[h!]
\centering
\includegraphics[width=0.85\textwidth]{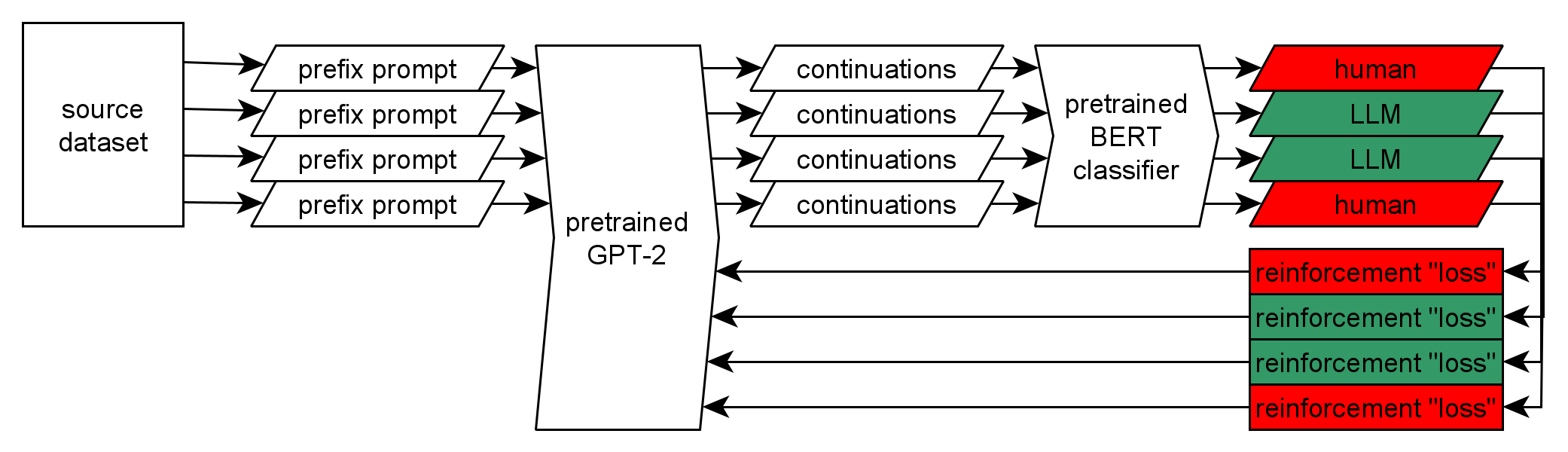}
\caption{GPT-2 training phase of the GAN for fake detection}
\label{gpt2_fake}
\end{figure}

\section{Training Plots from GPT-2 Fake Detection Training}
\label{fake_detection_training}

\subsection{training with fine-tuned GPT-2 on MS COCO dataset}
\label{fake_detection_training_fine_tuned}
 As we can see in Supplementary Fig. \ref{bert_acc}, the BERT model was not able to distinguish generated text from true text, even though there are small variations of one or two tokens for each sample.

Our hypothesis is that, while the loss decreases during training, BERT overfits the train set, and the features it uses to distinguish fake from true text do not generalize to the validation set.

\begin{figure}[h!]
\centering
\includegraphics[width=0.65\textwidth]{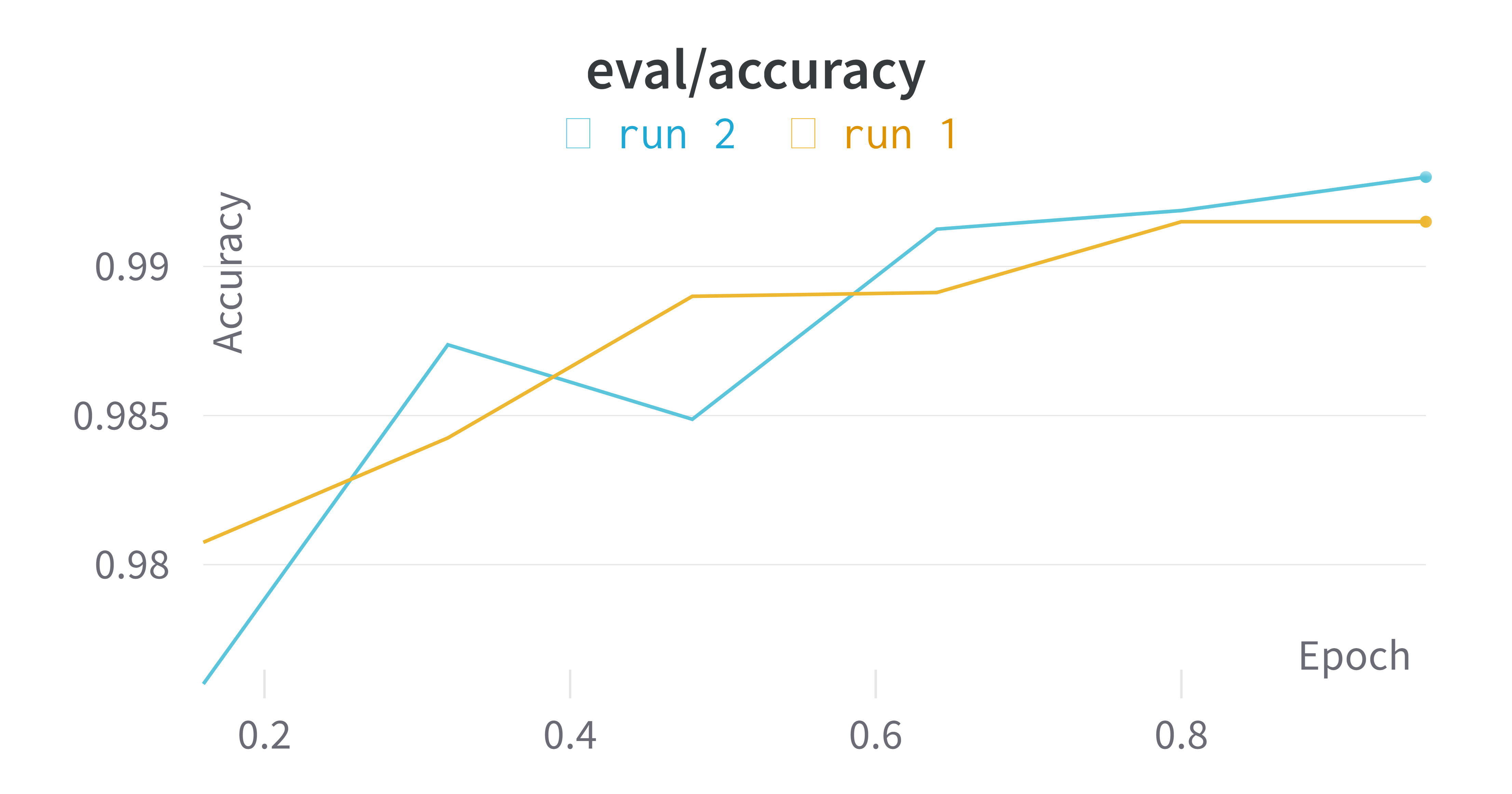}
\caption{BERT accuracy on the validation set when the samples from GPT-2 are not of the same length as the dataset on which BERT has been trained}
\label{bert_acc2}
\end{figure}

We also found while doing this experiment that, when the difference between the length of the fake samples and the true samples are different on average, BERT is able to pick up on that and have almost 100\% accuracy(cf. Supplementary Fig. \ref{bert_acc2}). The setup with GPT-2 and BERT seems tricky to train. Indeed, when GPT-2 outputs sampling resembling too much that of the dataset(variation of one of two tokens for example as for the above experiment), BERT is not able to distinguish true and fake data at all. When GPT-2 outputs samples that are too different from that of the dataset, BERT achieves almost 100\% accuracy. 

\subsection{training without fine-tuned GPT-2 on MS COCO dataset}

\label{fake_detection_training_no_fine_tune}
 In that scenario, BERT was able to distinguish with relatively high accuracy(about 90\% on average) fake from true samples(see Supplementary Fig. \ref{no_fine_tune}).
 
\begin{figure}[h!]
\centering
\includegraphics[width=0.65\textwidth]{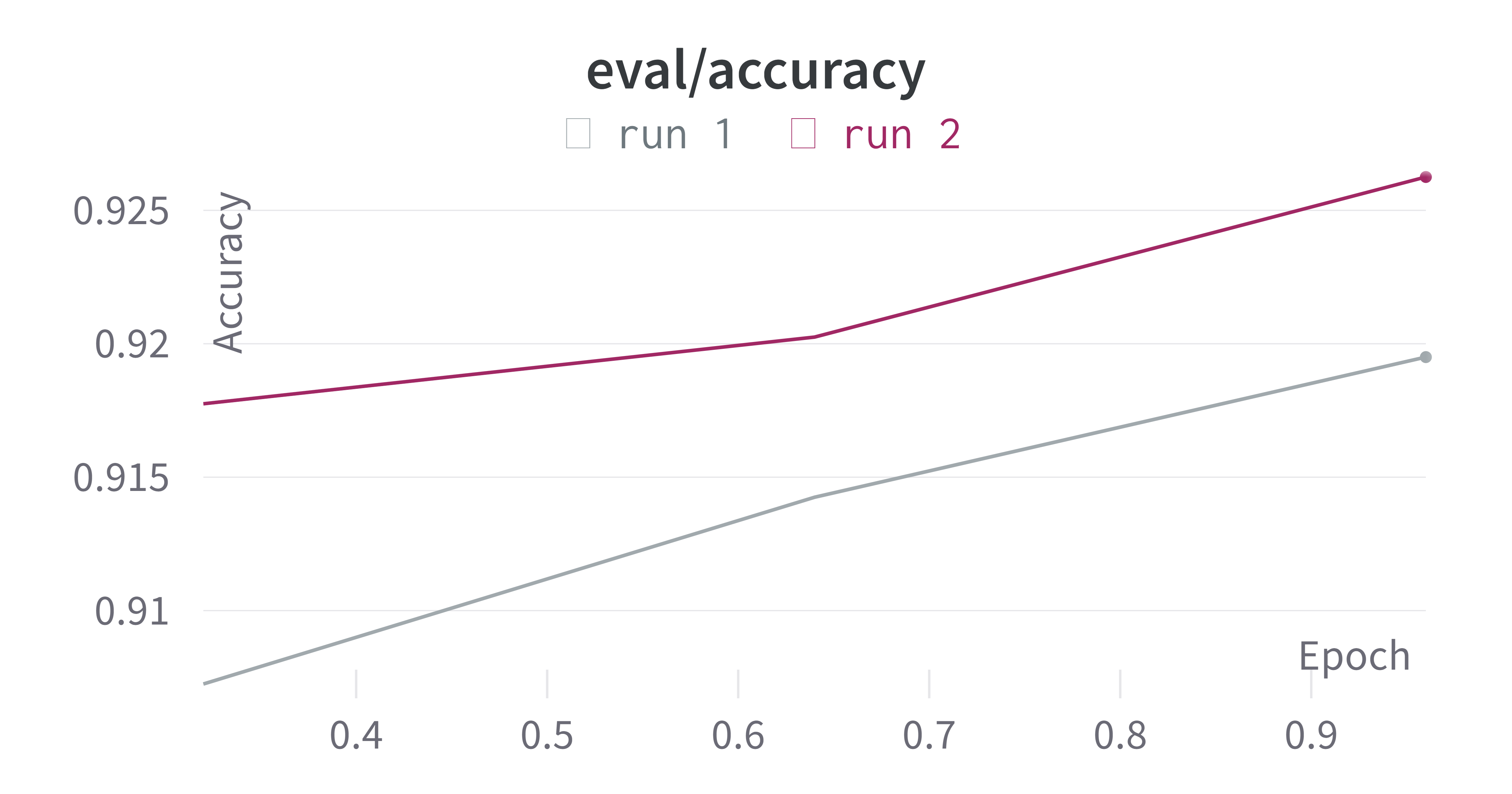}
\caption{BERT accuracy on the validation set when GPT-2 is not fine-tuned and length of GPT-2's output are adjusted to match the true dataset}
\label{no_fine_tune}
\end{figure}

However, GPT-2 was not able to train well. As we can see in Supplementary Fig. \ref{gpt2_fake_scores}, although in some epochs GPT-2 is able to produce more samples that fool BERT, in general, it is not able to and sometimes it gets worse(other runs gave similar results if not worse). Thus, in this setup, GPT-2 is not able to learn how to evade BERT fine-tuned for fake detection.

\begin{figure}[h!]
\centering
\includegraphics[width=0.65\textwidth]{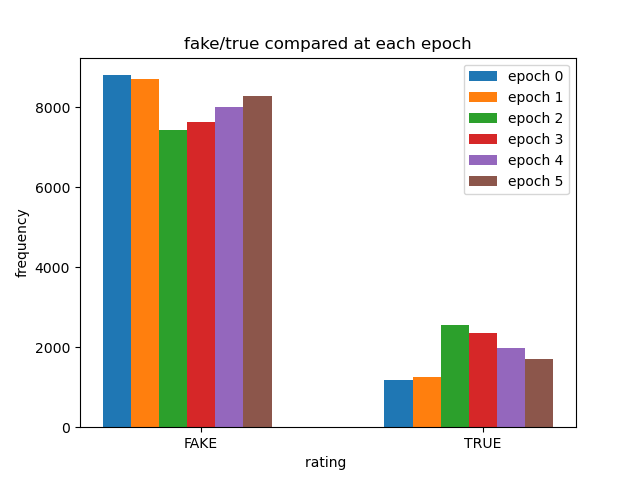}
\caption{GPT-2 training with scores from BERT fine-tuned for fake detection with MS COCO dataset}
\label{gpt2_fake_scores}
\end{figure}

\newpage

\subsection{training with partially fine-tuned GPT-2 on EMNLP news + MS COCO dataset}
 For the last epoch of training, about 40\% of samples generated by GPT-2, using prompts from the dataset, were classified as true (which we can compare with the 88\% accuracy of BERT for the validation dataset before adversarial training).
 
\begin{figure}[h!]
\centering
\includegraphics[width=0.65\textwidth]{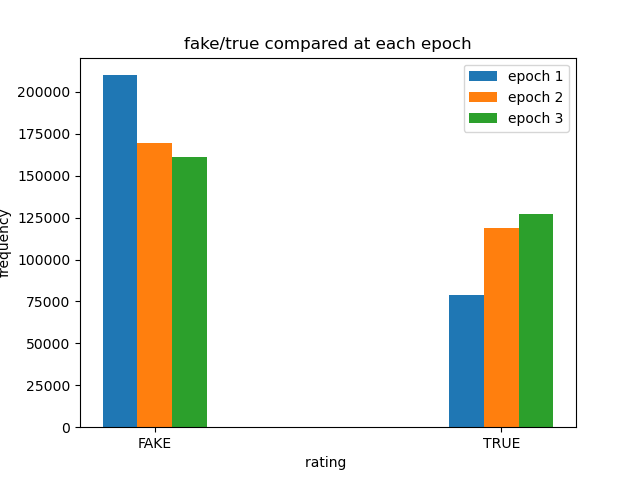}
\caption{GPT-2 training with scores from BERT fine-tuned for fake detection with EMNLP news dataset}
\label{gpt2_fake_scores_emnlp}
\end{figure}

\newpage

\section{Examples of Samples by GPT-2 during the Fake GAN Training and Score Given}
\label{score_samples}

Note: a score is a number between 0 and 1, which is computed by applying the softmax function to the probability of giving a positive score that BERT outputs given a sample as an input. This means that a score close to 1 means that BERT attributes a good probability that the sample is a true sample from the dataset, whereas a score close to 0 means that it assigns a low probability of being a true sample.

We also provided the corresponding sample in the dataset(created in the discriminator training phase) that has been given the label 1(true samples from a dataset) and the one that has been given the label 0(generated sample).
\begin{table}[!h]
	\centering
	\caption{samples generated by GPT-2 during fake GAN training and respective score}
	\small
	\begin{tabular}{*{1}{p{.98\linewidth}}}
		\toprule
		\textbf{bad score($< 0.5$)} \\\midrule
		\emph{"a bicycle replica with a clock as the base. The replica will be sold"} \textbf{score: 0.0057}\\
		\underline{label 0}:
		\emph{"a bicycle replica with a clock as the baseplate. Another"}\\
		\underline{label 1}: \emph{"a bicycle replica with a clock as the front wheel."}\\\midrule
		\emph{"a car that seems to be parked illegally in a Houston suburb. A car"} \textbf{score: 0.0057}\\
		\underline{label 0}:
		\emph{"a car that seems to be parked illegally. A young"}\\
		\underline{label 1}: \emph{"a car that seems to be parked illegally behind a legally parked car"}\\\midrule
		\emph{"a black honda motorcycle parked in front of a hotel in Seoul, South Korea"} \textbf{score: 0.0633}\\
		\underline{label 0}:
		\emph{"a black honda motorcycle parked in front of the garage"}\\
		\underline{label 1}: \emph{"a black honda motorcycle parked in front of a garage."}\\\midrule

		\textbf{good score($\geq 0.5$)} \\\midrule
		\emph{"a car that seems to be parked illegally near the Trump Tower in New York"} \textbf{score: 0.794}\\
		\underline{label 0}:
		\emph{"a car that seems to be parked illegally. A young"}\\
		\underline{label 1}: \emph{"a car that seems to be parked illegally behind a legally parked car"}\\\midrule
		\emph{"a honda motorcycle parked in a grassy area in this file photo in Oakland"} \textbf{score: 0.976}\\
		\underline{lable 0}:
		\emph{"a honda motorcycle parked in a grassy area near the"}\\
		\underline{label 1}: \emph{"true: a honda motorcycle parked in a grass driveway"}\\
		\bottomrule
	\end{tabular}
\end{table}

\newpage
\section{Example of Samples Generated by GPT-2 during Fake GAN Training with EMNLP News + MS COCO Dataset after one Epoch of Adversarial Training}
\label{samples_emnlp_coco}
\begin{table}[!h]
	\centering

	\label{samples_gpt2_fake_emnlp}
	\small
	\begin{tabular}{*{1}{p{.98\linewidth}}}
		\toprule
		a bicycle replica with a clock as the:: the at:: into:::, in as a -: for: around \textbackslash xad: it at of said more said: ' on the: the the:: reportedly said,:- police, \textbackslash xad: at: the said,\\
		a black honda motorcycle parked in front for on an said: the said: \textbackslash n, said said,,,:, but: as -, a in said the not a. said:: have had, in had in.: and: at, -: " in -,'\\
		a room with blue walls and a white::::: said \textbackslash xad: on the, have it: said,:: after  \textbackslash n: -,. more::.: in in, in: not:, \textbackslash n to said at  \textbackslash xad:: of.:. said:\\

		\bottomrule
	\end{tabular}
\end{table}

\end{document}